# Blockchain-Based Federated Learning: Incentivizing Data Sharing and Penalizing Dishonest Behavior

Amir Jaberzadeh[1], Ajay Kumar Shrestha[2][0000-0002-1081-7036], Faijan Ahamad Khan[1], Mohammed Afaan Shaikh[1], Bhargav Dave[1] and Jason Geng[1]

[1] Bayes Solutions, 840 Apollo St, El Segundo CA 90245, USA
`jason@bayes.global`
[2] Vancouver Island University, 900 Fifth St, Nanaimo, BC V9R 5S5, Canada
`ajay.shrestha@viu.ca`

**Abstract.** With the increasing importance of data sharing for collaboration and innovation, it is becoming more important to ensure that data is managed and shared in a secure and trustworthy manner. Data governance is a common approach to managing data, but it faces many challenges such as data silos, data consistency, privacy, security, and access control. To address these challenges, this paper proposes a comprehensive framework that integrates data trust in federated learning with InterPlanetary File System, blockchain, and smart contracts to facilitate secure and mutually beneficial data sharing while providing incentives, access control mechanisms, and penalizing any dishonest behavior. The experimental results demonstrate that the proposed model is effective in improving the accuracy of federated learning models while ensuring the security and fairness of the data-sharing process. The research paper also presents a decentralized federated learning platform that successfully trained a CNN model on the MNIST dataset using blockchain technology. The platform enables multiple workers to train the model simultaneously while maintaining data privacy and security. The decentralized architecture and use of blockchain technology allow for efficient communication and coordination between workers. This platform has the potential to facilitate decentralized machine learning and support privacy-preserving collaboration in various domains.

## 1   Introduction

In recent years, data sharing has become increasingly important for collaboration and innovation in various fields. The adoption of secure and trustworthy multi-center machine learning poses numerous challenges, including data sharing, training algorithms, storage, incentive mechanisms, and encryption. In this paper, we aim to tackle these challenges and propose a comprehensive solution for collaborative machine-learning applications. However, managing and sharing data in a secure and trustworthy manner poses several challenges, such as data silos, privacy, security, access control, and data consistency. Data governance has been proposed as a common approach to managing data, but it still faces several challenges, and as a result, data trust has emerged as a nascent sub-area of data management [1].

This research paper proposes a trustworthy and robust framework for federated learning participants. Federated Learning (FL) is a privacy-preserving distributed Machine Learning (ML) paradigm [2]. The proposed comprehensive framework integrates data trust, InterPlanetary File System[1] (IPFS), blockchain, and smart contracts to establish a secure and mutually beneficial data-sharing distributed FL platform. The framework is designed to provide incentives, access control mechanisms, and penalties for any dishonest or malicious behavior — sharing bad data or non-compliance with protocols. The framework aims to foster trust among stakeholders, encourage data sharing for mutual benefit, and discourage actions that may compromise data security and accuracy. Our proposed approach is built on the use of smart contracts that enable monitoring of data sharing, access control, and compensation. To participate in the federated learning process, users must register and contribute their data while being required to provide a collateral deposit to deter dishonest behavior.

Our proposed framework prioritizes data privacy and security by utilizing the encryption-enabled InterPlanetary File System (IPFS) as a decentralized peer-to-peer file system to store and access data. By using IPFS, federated learning models can be trained on data that is stored on a distributed network of users' devices, reducing the need for centralized storage. The utilization of encryption enabled IPFS ensures that the user's data privacy is safeguarded throughout the learning process. The framework is designed to provide a fair and transparent approach to compensate users for their contributions while ensuring the privacy and security of their data.

The rest of the paper is organized as follows. Section 2 provides a succinct analysis of existing architectures and identifies their shortcomings. Our proposed model for the solution architecture is presented in Section 3. Section 4 contains the experimental results and discussion. Lastly, Section 5 concludes the paper by outlining future directions for further research and improvements to the proposed model.

## 2    Background and Related Works

The primary challenge in data governance is to dismantle data silos [3] and ensure data consistency, compatibility, privacy, security, access control, ownership, and rewards for sharing. It is, therefore, imperative to have data governance frameworks that can evolve with new technologies to address emerging challenges. FL is a learning technique where a central server connects with many clients (e.g., mobile phones, and pads) who keep their data private. Since communication between the central server and clients can be a bottleneck, decentralized federated learning (DFL) [4] is used to connect all clients with an undirected graph, which reduces communication costs and increases privacy by replacing server-client communication with peer-to-peer communication. DFL offers communication efficiency and fast convergence, and the advantages of FL are summarized in [5].

---

[1] https://ipfs.tech/

Several variants of Federated Average (FedAvg) [2] exist with theoretical guarantees. In [6], the momentum method is used for local client training, while [7] proposes adaptive FedAvg with an adaptive learning rate. Lazy and quantized gradients are used in [8] to reduce communications, and in [9], the authors propose a Newton-type scheme. Decentralized (sub) gradient descents (DGD) are studied in [10-13], and DSGD is proposed in [14]. Asynchronous DSGD is analyzed in [15], and quantized DSGD is proposed in [16]. Decentralized FL is popular when edge devices do not trust central servers to protect their privacy [17]. Finally, the authors in [16] propose a novel FL framework without a central server for medical applications. The authors in [18] propose a secure architecture for privacy-preserving in smart healthcare using Blockchain and Federated Learning, where Blockchain-based IoT cloud platforms are used for security and privacy, and Federated Learning technology is adopted for scalable machine learning applications. In [19], authors propose a blockchain-based Federated Learning (FL) scheme for Internet of Vehicles (IoV), addressing security and privacy concerns by leveraging blockchain and a reputation-based incentive mechanism.

Compared to prior research on FedAvg, our paper proposes a decentralized framework to improve FL's resilience to node failures and privacy attacks. Unlike previous decentralized training approaches, our algorithm utilizes IPFS and efficient encryption methods to securely converge model training among many nodes.

## 3 System Model

### 3.1 Data Trust, Access Control and Incentive Method

Data trust ensures that data is available for data mining with increasing legal protections for privacy and sustains the underlying ownership of the data and digital rights, which is the primary focus of the data management field [20]. Our work emphasizes sharing data, transparency, control, and incentives for users in the federated learning setting. Specifically, it explores a particular type of technical platform, distributed ledgers with smart contracts. The smart contract is designed to oversee data sharing, compensation, and access control. Participants would be allowed to register and contribute their data to the federated learning process, and a collateral deposit would be required from each participant to discourage any dishonest behavior. The collateral deposit serves as a financial penalty for participants who fail to provide quality data or who intentionally provide misleading information. If a participant fails to provide accurate data or engages in any dishonest behavior, the deposit will be forfeited. The forfeited deposit will then be used to compensate other participants who have contributed accurate data to the federated learning process. Through the implementation of a smart contract, the total compensation for data sharing is updated and distributed to participants based on their contribution. The contract would also ensure that each participant can only register once, and that compensation can only be distributed when the total compensation amount is positive. The proposed smart contract system provides a reliable and secure framework for federated learning, user data, and blockchain integration. It offers a fair and transparent way of compensating participants for their contributions while ensuring the privacy and security of the data.

### 3.2 IPFS Storage

In addition to privacy concerns, data storage is another key challenge in federated learning. Traditional data storage approaches are not well-suited for federated learning, as they often require centralized storage of data. This centralized storage approach can increase the risk of data breaches and raises concerns around data ownership and control. To address this challenge, we have proposed using the InterPlanetary File System (IPFS), a peer-to-peer distributed file system that allows data to be stored and accessed in a decentralized manner. By using IPFS, federated learning models can be trained on data that is stored on users' devices, without the need for centralized storage.

### 3.3 Confidentiality and Privacy

In the context of IPFS, hashing is not sufficient to ensure the confidentiality and privacy of the stored data, as the content can still be accessed by anyone who has access to the network. To address this, both symmetric key encryption and asymmetric cryptography are applied [21], and we further use smart contracts for access control and to provide confidentiality and privacy to the data stored in IPFS. This approach ensures that even if an attacker gains access to the IPFS network, they would not be able to read the encrypted content without the secret. Although encryption may introduce some overhead in terms of performance and complexity, it is necessary for ensuring the security of data in IPFS.

We used the cryptography Python library to securely share and store machine learning models using IPFS. The library provides various cryptographic primitives and recipes for encryption, digital signatures, key derivation, hashing, and more, adhering to best security practices. The code initially connects to an IPFS daemon, loads a model from IPFS, generates an RSA key pair and an AES key, and encrypts the AES key with the public key using hybrid encryption. The actual data is encrypted using the symmetric key (AES), and the symmetric key is encrypted using an asymmetric key (RSA). This ensures that the data can only be decrypted by the intended recipient who possesses the corresponding private key. Our research further implements a method for encrypted model states to be fetched from a group of workers, decrypted with the AES key, and returned as decrypted model states. This allows multiple parties to share their model states securely. In addition, we have also implemented a method for pushing a model state to IPFS. The model state is encrypted using the AES key and the AES key is encrypted with the public key. This encryption mechanism allows the model state to be stored on the IPFS network in an encrypted form, ensuring that only authorized parties can access it. To optimize memory usage, our code maintains a list of model hashes that have been pushed to IPFS and clears the list once a specified number of models have been pushed. This optimization technique helps prevent system resources from being overwhelmed and causing performance issues. By clearing the list after a certain number of models have been pushed, the code ensures that memory usage remains within reasonable limits.

### 3.4 Decentralized Network Architecture

Our research proposes a blockchain-based architecture for federated learning, consisting of a smart contract and IPFS. The smart contract coordinates the FL task, distributing rewards and penalizing bad actors. The metric used for rewarding workers based on performance considers factors such as model accuracy, consistency, precision and recall on the unseen test dataset. Our proposed architecture contains essential information such as the participants' details, model evaluations submitted at the end of each round, and the reward to be distributed. The IPFS, on the other hand, stores the models trained by participants at the end of each round.

As shown in Figure 1, blockchain-based federated learning involves two classes of actors: the requester and the workers. The requester initiates the FL task by deploying the smart contract, pushing the initial model to the IPFS, and specifying additional parameters such as the number of rounds and the reward to be distributed. The requester can push any model for any task to IPFS. On the other hand, workers participate in the FL task created by the requester. They train the model through a round-based system on their own data and earn rewards based on their performance.

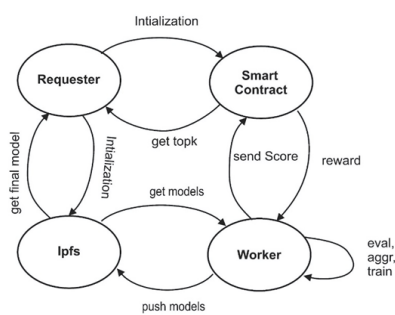

**Fig. 1.** Blockchain and IPFS-based federated learning.

**Workflow.** To begin the task, the requester deploys the smart contract and sets the number of training rounds (N) and the total reward (D) for workers. The requester also pushes the initial model to the IPFS-based model storage, which will be used as the basis for the trained models. Workers can then join the task by interacting with the smart contract, and once enough workers have joined, the requester triggers the training phase through the smart contract. During the training phase, workers train the model for N rounds on their local data, with each round beginning with workers retrieving the trained models of all other workers from the IPFS-based model storage to evaluate them on their local data. The scores are pushed to the smart contract, which aggregates them to obtain the Top K best-performing workers in the previous round. Rewards are distributed to the workers based on their performance. The trained models are then pushed to the IPFS-based model storage, and the process repeats for N rounds. The requester is not involved in any interaction during the training phase, but some operations such as score aggregation can be offloaded to the requester machine to save on computational resources and transaction costs. Once the training phase concludes, the requester can retrieve the final global model from the IPFS-based model storage and close the task by calling a function of the smart contract.

**Smart Contracts.** The smart contract contains various functions that enable the task requester to initialize, initiate, and oversee the FL task. It also allows workers to participate in the task, submit evaluations, and exit from it. Below is a brief overview of the different functions within the smart contract.

*initializeTask*. This function is called by the requester to initialize the FL task. It takes two parameters: the URI of the machine learning model and the number of rounds in the FL task. The function requires a deposit to be made in the smart contract.

*startTask*. This function is called by the requester to start the FL task. It changes the status of the task to "running".

*joinTask*. This function is called by the workers to join the FL task. It registers the worker in the smart contract and returns the URI of the machine learning model.

*submitScore*. This function is called by the workers to submit the score of their local model after each round's evaluation phase.

*removeWorker*. This function is called by workers to remove themselves from the task.

*nextRound*. This function is called by the requester to advance the FL task to the next round.

*getSubmissions*. This function is called by the requester to get the submissions from all workers for the current round.

*submitRoundTopK*. This function is used to get the top k rank of the workers who will be rewarded for their performance in a task or job. This information is used to distribute the rewards among the top-performing workers.

*distributeRewards*. This function is used to reward top-performing users in a round by splitting the total reward among them. The first users receive half of the total reward, while the remaining users receive a smaller share.

### 3.5 Aggregation/Averaging method

As shown in Figure 2, in federated learning, workers train the model on their local data, and their models are stored in IPFS-based model storage. The workers retrieve their own model and the models of other workers from the storage and append them to a dictionary. To improve the model's accuracy, the workers use an averaging function that takes all the stored models as input and returns the average model. The averaging is done by adding up all the models and dividing the result by the number of workers who contributed to their models.

Overall, this process allows for multiple workers to collaborate on model training without relying on centralized data storage. By averaging the models, the final global model can be improved by incorporating the knowledge of all workers in the task. By contrast, FedAvg in centralized fashions leads to massive communication between the central server and clients, which causes channel blocking. Furthermore, the central server is vulnerable to attack, compromising the privacy of the entire system.

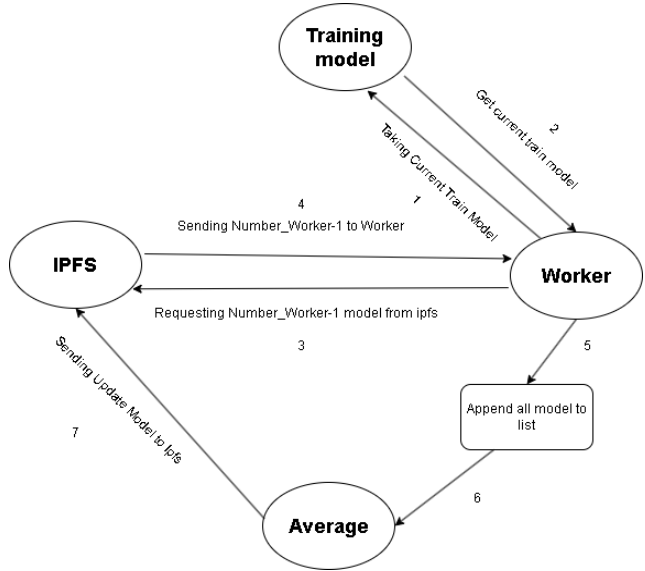

**Fig. 2.** Aggregation/Averaging methods in our DFL framework.

## 4      Experimental Results and Discussions

We evaluated the proposed platform using the MNIST dataset[2]. We used a simple feed-forward Neural Network (CNN) model with N layers to classify 0-9 handwritten numbers. The algorithms were developed using the PyTorch framework on the Ethereum blockchain. The simulation modeling was done on Ganache, a local Ethereum blockchain for testing. The training dataset comprised 60,000 images, while the test dataset consisted of 10,000 images. The training dataset was divided evenly between workers at the start of the training and each worker used the test dataset for their evaluations and scoring. Our implementation employed a decentralized client and server system and ran on a local machine. When the requester initiated the process, each worker trained the model taken from IPFS sequentially and securely saved the trained model to the IPFS until the maximum number of epochs was reached. Alternatively, workers could be spawned in parallel and train the model simultaneously on multiple devices. Here we tested our results with one machine that runs each step sequentially.

### 4.1      Performance Analysis

We first trained the classification model with our Decentralized Federated Learning framework and showed a convergence of above 95% accuracy under 90 epochs as shown in Figure 3. We also studied other performance metrics like precision and recall and got 0.973 and 0.97 respectively which shows great performance in the classification task. The total training time by 3 workers was 6525.46 seconds. Each worker takes about 36 minutes to converge on a Xeon CPU with 8 cores which is a comparable time to convergence in decentralized Federated Learning frameworks. We also compared the impact of double encryption on the convergence time. As shown in the left graph of Figure 3, there is an additional 2 minutes and 34 seconds overhead for all three workers or 51 seconds for each worker. The communication cost for our double encryption and decryption process and secure key pair transfer protocol was only 2% of the time required for convergence with the same accuracy. The accuracy for each worker is plotted sequentially for each round of 3 epochs. As shown in Figure 3, when the models start to train, all three workers start with low accuracies that improve within their first round (i.e., 3 epochs) and followed by the next worker's training.

---

[2] https://pytorch.org/vision/stable/generated/torchvision.datasets.EMNIST.html#torchvision.datasets.EMNIST

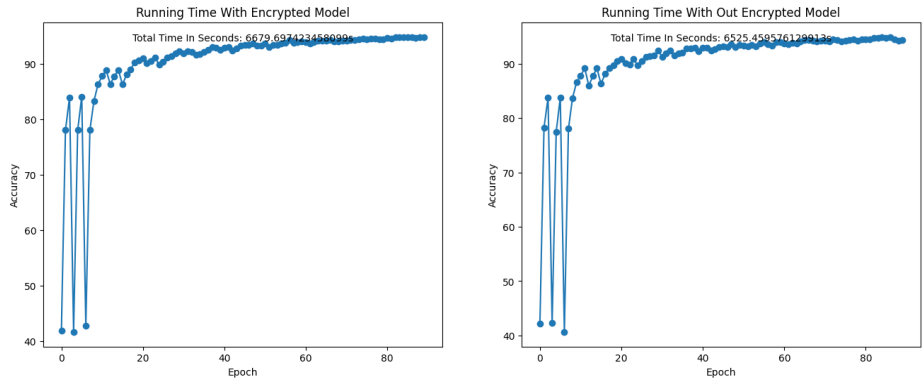

**Fig. 3.** Comparison of Running Time: Encrypted vs Unencrypted Models for 3-Worker Model.

### 4.2 Accuracy (Workers vs Epochs) Analysis

The graphs in Figure 4 show the accuracy of a federated learning model trained over multiple epochs, with the left graph showing results for a model trained with 3 workers and the second graph showing results for a model trained with 5 workers. Accuracy is the percentage of correct classifications that a trained machine learning model achieves. By analyzing the graphs, we can see that both models reach an acceptable accuracy over a similar number of epochs.

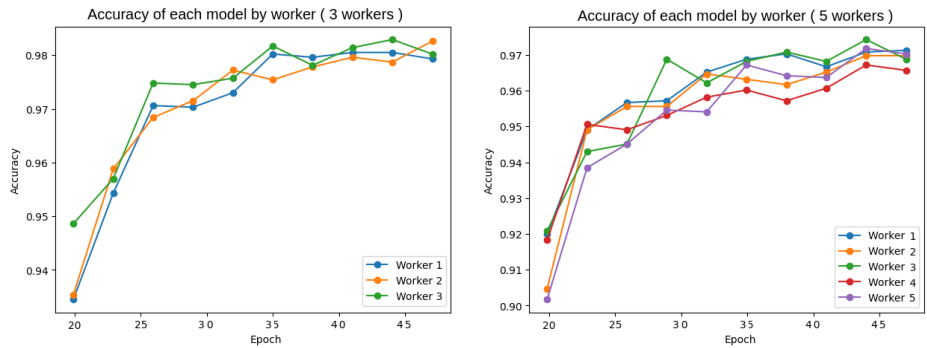

**Fig. 4.** Accuracy vs Epochs for 3-worker (left) and 5-worker (right) models.

This shows that dividing data between more workers does not have a negative impact on model convergence but can speed up the training process and scale up the training process. It can also reduce the required compute power for each worker which qualifies low-end devices to be used as compute nodes. In the left graph with 3 workers, the model's accuracy has a more stable pattern, which is due to having more data to train on for each worker. In a realistic case by increasing the training dataset, we can improve the stability of models with more workers. Overall, we can conclude that the decentralized federated learning model is performing well and improving over time, but the number of workers should be chosen as a ratio of the training dataset.

## 5     Conclusion

We proposed a decentralized federated learning architecture that leverages blockchain, smart contracts and IPFS for secure and efficient training of a global model with decentralized data. Experimental results showed that our proposed framework achieved above 95% accuracy under 90 epochs with a comparable convergence time to centralized federated learning frameworks. We also compared the impact of double encryption on the convergence time and showed that it only resulted in a minimal overhead cost of 2%. Overall, our proposed approach addresses several challenges associated with managing and sharing data in a secure and trustworthy manner by providing a comprehensive framework that establishes trust among stakeholders, promotes data sharing that benefits all involved parties, and deters any actions that could compromise the security and accuracy of the shared data. Future research can explore the scalability and feasibility of the proposed model in a real-world scenario.

## References

1. Delacroix, S. & Lawrence, N. D.: Bottom-up data Trusts: Disturbing the 'one size fits all' approach to data governance. *International Data Privacy Law* **9**, 236–252 (2019)
2. Brendan McMahan, H., Moore, E., Ramage, D., Hampson, S. & Agüera y Arcas, B.: Communication-efficient learning of deep networks from decentralized data. in *Proceedings of the 20th International Conference on Artificial Intelligence and Statistics, AISTATS 2017* (2017)
3. Seaman, D.: From isolation to integration: re-shaping the serials data silos. *Serials: The Journal for the Serials Community* **16**, 131–135 (2003)
4. Zinkevich, M. A., Weimer, M., Smola, A. & Li, L.: Parallelized stochastic gradient descent. in *Advances in Neural Information Processing Systems 23: 24th Annual Conference on Neural Information Processing Systems 2010, NIPS 2010* (2010)
5. Sun, T., Li, D. & Wang, B.: Decentralized Federated Averaging. *IEEE Trans Pattern Anal Mach Intell* (2022). doi:10.1109/TPAMI.2022.3196503
6. Hsu, T.-M. H., Qi, H. & Brown, M.: Measuring the Effects of Non-Identical Data Distribution for Federated Visual Classification (2019)

7. Deng, Y., Kamani, M. M. & Mahdavi, M.: Adaptive Personalized Federated Learning (2020)
8. Sun, J., Chen, T., Giannakis, G. B. & Yang, Z.: Communication-Efficient Distributed Learning via Lazily Aggregated Quantized Gradients. *Advances in Neural Information Processing Systems* **32** (2019)
9. Li, T. *et al.*: FedDANE: A Federated Newton-Type Method. in *2019 53rd Asilomar Conference on Signals, Systems, and Computers* 1227–1231 (IEEE, 2019). doi:10.1109/IEEECONF44664.2019.9049023
10. Nedić, A. & Ozdaglar, A.: Distributed subgradient methods for multi-agent optimization. *IEEE Trans Automat Contr* **54**, 48–61 (2009)
11. Chen, A. I. & Ozdaglar, A.: A fast distributed proximal-gradient method. in *2012 50th Annual Allerton Conference on Communication, Control, and Computing, Allerton 2012* 601–608 (2012). doi:10.1109/Allerton.2012.6483273
12. Jakovetic, D., Xavier, J. & Moura, J. M. F.: Fast distributed gradient methods. *IEEE Trans Automat Contr* **59**, 1131–1146 (2014)
13. Matei, I. & Baras, J. S.: Performance evaluation of the consensus-based distributed subgradient method under random communication topologies. *IEEE Journal on Selected Topics in Signal Processing* **5**, 754–771 (2011)
14. Lan, G., Lee, S. & Zhou, Y.: Communication-efficient algorithms for decentralized and stochastic optimization. *Mathematical Programming* **180**, 237–284 (2020)
15. Lian, X., Zhang, W., Zhang, C. & Liu, J.: Asynchronous decentralized parallel stochastic gradient descent. in *35th International Conference on Machine Learning, ICML 2018* vol. 7 4745–4767 (2018)
16. Reisizadeh, A., Mokhtari, A., Hassani, H. & Pedarsani, R.: Quantized Decentralized Consensus Optimization. in *Proceedings of the IEEE Conference on Decision and Control* vols 2018-Decem 5838–5843 (2018)
17. Sun, T. *et al.*: Non-ergodic convergence analysis of heavy-ball algorithms. in *33rd AAAI Conference on Artificial Intelligence, AAAI 2019, 31st Innovative Applications of Artificial Intelligence Conference, IAAI 2019 and the 9th AAAI Symposium on Educational Advances in Artificial Intelligence, EAAI 2019* (2019). doi:10.1609/aaai.v33i01.33015033
18. Singh, S., Rathore, S., Alfarraj, O., Tolba, A. & Yoon, B.:A framework for privacy-preservation of IoT healthcare data using Federated Learning and blockchain technology. *Future Generation Computer Systems* **129**, 380–388 (2022)
19. Wang, N. et al.: A blockchain based privacy-preserving federated learning scheme for Internet of Vehicles. *Digital Communications and Networks* (2022). doi:10.1016/j.dcan.2022.05.020
20. Shrestha, A. K., Vassileva, J. & Deters, R.: A Blockchain Platform for User Data Sharing Ensuring User Control and Incentives. *Frontiers in Blockchain* **3**, 48 (2020)
21. Shrestha, A. K.: Designing Incentives Enabled Decentralized User Data Sharing Framework. *UofS Harvest* (2021). https://hdl.handle.net/10388/13739